%% file: main.tex
\documentclass[10pt,twocolumn,letterpaper]{article}
\usepackage[pagenumbers]{cvpr} 

\usepackage{graphicx}
\usepackage{amsmath}
\usepackage{amssymb}
\usepackage{booktabs}
\usepackage{multirow}
\usepackage[accsupp]{axessibility}
\usepackage[pagebackref,breaklinks,colorlinks]{hyperref}

\usepackage[capitalize]{cleveref}
\crefname{section}{Sec.}{Secs.}
\Crefname{section}{Section}{Sections}
\Crefname{table}{Table}{Tables}
\crefname{table}{Tab.}{Tabs.}

\begin{document}

	\title{Leveraging Multi-view Data for Improved Detection Performance: \\ An Industrial Use Case}
	
	\author{Faranak Shamsafar\hspace{0.5cm} Sunil Jaiswal\hspace{0.5cm} Benjamin Kelkel\hspace{0.5cm} Kireeti Bodduna\hspace{0.5cm} Klaus Illgner-Fehns\\
		K\textbar Lens GmbH, Germany\\
	}
	\maketitle
	\footnotetext{\hspace{-0.5cm}Corresponding author: {\tt\small {faranak.shamsafar}@k-lens.de} \\
		This work was partially funded by the German Ministry for Education and Research (BMBF) under the grant PLIMASC.}
	
	\begin{abstract}
		Printed circuit boards (PCBs) are essential components of electronic devices, and ensuring their quality is crucial in their production. However, the vast variety of components and PCBs manufactured by different companies makes it challenging to adapt to production lines with speed demands. To address this challenge, we present a multi-view object detection framework that offers a fast and precise solution. We introduce a novel multi-view dataset with semi-automatic ground-truth data, which results in significant labeling resource savings. Labeling PCB boards for object detection is a challenging task due to the high density of components and the small size of the objects, which makes it difficult to identify and label them accurately. By training an object detector model with multi-view data, we achieve improved performance over single-view images. To further enhance the accuracy, we develop a multi-view inference method that aggregates results from different viewpoints. Our experiments demonstrate a 15\% improvement in mAP for detecting components that range in size from 0.5 to 27.0 $mm$. 
	\end{abstract}
	
	\section{Introduction}
	Every electronic device, including personal computers, televisions, or advanced industrial, transportation equipment, has a printed circuit board (PCB). The PCB is ultimately responsible for the functionality of the whole system, so ensuring PCB quality in a timely and precise manner is essential. In industrial environments, human-operated and manual PCB analysis are inefficient, expensive, and suffer from a high error rate and subjective accuracy. This makes the \emph{automatic} inspection of PCBs essential, which is accomplished primarily through visual analysis.
	
	Generally, PCB inspection can be carried out by using different imaging modalities, \ie gray-scale/RGB images \cite{lu2020fics,zhao2022pcb}, depth maps \cite{herchenbach2013segmentation,li2020semantic}, or infrared images \cite{li2021multisensor, huang2005vq}. Although RGB data is challenging due to the variety of illuminations, reflections, and colors of components among similar types, the images under visible light present rich data for PCB analysis. Nevertheless, PCB design sets vary tremendously based on their functionality, manufacturer, and component variations and hence, more complexity exists in PCB images compared to images in other common computer vision scenarios.
	
	Referencing a PCB is a shallow approach for PCB inspection, in which the test image of the PCB must strictly match the reference image of the same PCB. Alternatively, in earlier works, indirect abstract information was derived from PCB images, such as the number of objects \cite{wu1996automated}, the number of connected holes in the PCB \cite{tatibana1997novel}, or solder joint locations \cite{luo2007ann}. On the basis of classical image processing techniques, \cite{crispin2007automated} used the normalized cross correlation template-matching and proposed a method for constraining the genetic algorithm search space to reduce computational calculations. Authors in \cite{zeng2011algorithm} used color distribution patterns for recognizing the components. Similarly, in \cite{li2013smd}, authors proposed a color distribution-based segmentation for resistors and integrated circuits (IC). Yet, all these referential techniques lead to PCB-specific solutions.
	
	Typically, PCB inspection based on component placement is the most reliable approach that can be exploited in various PCB-related applications, including: 1) PCB fabrication and assembly in PCB manufacturers, 2) Optimization of PCB component placement, 3) Checking for malicious inputs to the PCBs for security issues, 4) PCB recycling: Recovering the reusable components and precious materials and separating the toxic substances, and 5) Quality control in PCB manufacturers or in electronics producers (before PCBs are assembled into electronic devices), \eg locating the missing or misaligned components.

	After computer vision witnessed its renaissance via deep learning, different approaches emerged for automated PCB inspection. A two-stage approach was proposed in \cite{lim2019smd} that firstly extracted regions and then applied a classification using a convolutional neural network (CNN). \cite{lian2021automatic} integrated a geometric attention-guided mask branch into an object detector for IC detection. In \cite{lu2020fics}, the capability of classification on the component images of six classes was explored. In \cite{kuo2019data}, a multiple-stage approach is proposed in which a class-agnostic region proposal network is followed by a low-shot similarity prediction classifier and a graph-based neural network to refine features, which showed poor performance for small components. In these deep learning-based solutions providing suitable and adequate images still remains an open issue. Obtaining high-quality ground-truth annotations is another level of challenge for deep learning-based PCB analysis. 
	\input{figBlockgiagram}
	
	In this work, we address both \emph{image modality} and \emph{image analysis} for PCB inspection and propose an approach that processes the whole image in an end-to-end framework, targeted to detect and identify the components of a PCB with high accuracy. Evidently, the creation of a dataset for deep learning-based PCB object detection is highly expensive and time-consuming. Therefore, we develop a framework for hardware-augmented multi-view training data generation that includes semi-automatic labeling, making the annotation process much smoother with consistent labels. Using this multi-view dataset, we gain a significant improvement in the model performance. Moreover, a multi-view inference module is formulated to accumulate the results from different viewpoints and reach a consensus for more accurate and reliable performance. Our framework can effectively inspect the PCBs with component dimensions as small as 0.5 $mm$. In addition, the approach is applicable to inspection of different PCB samples in open-ended and outsourced PCB collections and does not rely on a specific type or a reference sample. An overview of our model is shown in \cref{fig:Blockgiagram}. 
	\section{Methodology}\label{sec:Methodology}
	\subsection{Hardware setup}
	To collect images, we used the 3rd generation K\textbar Lens (80 $mm$ focal length) mounted on a 61 MP full-frame camera (hr455CXGE, SVS-VISTEK) (\cref{fig:a}). K\textbar Lens \cite{klens, klensOptics, garcia2023cnn} via its special technology records multi-view images and these images can be used in its software to compute disparity maps. Multi-view images are formed in a $3\times3$ grid of images, named kaleidoscopic image. The technology involves refracting light rays from the main lens onto an intermediate image plane via a mirror system. The rays are then projected onto the sensor in a way that they are split into nine separate images. This is similar to simultaneous shooting by nine cameras placed at very short distances from each other. However, compared to a multi-array camera setup, calibration and rectification are well-developed and stable in K\textbar Lens. This setup is also highly compact for capturing multi-view images, which is particularly useful for tiny and close-range objects, \ie macro photography. 
	
	For a larger magnification, a close-up lens with 5 dioptres (MARUMI) was attached to the lens. The PCBs were illuminated in a bright field setup with a flat top light (DTL 1010 WTS, MBJ) with white LEDs. The light source was further enhanced with a polarization filter to reduce reflections from pins. The working distance was set to around 100 $mm$.
	\input{fighardwaresetup}	
	\subsection{K\textbar Lens PCB dataset}	
	Our goal is not to limit the dataset to a specific manufacturer's PCBs or a particular PCB. This attitude introduces wide variance in PCBs in terms of design, component types, color, \etc. It may be possible to generate datasets synthetically. However, this approach requires CAD models, is PCB-specific, and cannot represent real-world data distribution. Moreover, the manual labeling effort for these images remains a challenge. 
	
	In this context, we create a PCB dataset with the following data types: 1) High-resolution RGB images, 2) High precision depth maps to the accuracy of about 70 $\mu m$, and 3) Multi-view RGB samples (nine images) obtained in each image shooting. We leverage the merit of having all these data types in one shot to analyze PCBs using deep learning.
	
	Five PCBs with different functionalities are used for image collection. The lateral dimensions of the PCB components vary from $0.5\times1.0$ to $27.0\times 27.0$ $mm^2$. In each shot, a region of about $38 \times 56\:mm^2$ from the PCBs is recorded in a scanning manner with an approximate overlap of $15$ to $25\:mm$. Overall, a total of 262 kaleidoscopic images ($3\times3$ grid multi-view) are captured (sample shown in \cref{fig:b}). 
	
	To extract the individual images, the software provided by the K\textbar Lens system mirrors, rectifies, and crops the viewpoints (\cref{fig:c}). Note that the individual images record the scene from slightly different perspectives, from which the center viewpoint presents the least distortion and is referred to as single-view hereon. We also record the disparity maps, which show the displacements of image content in a pair of rectified images, namely the displacements between the center-view image and every other viewpoint. A sample disparity map from the center viewpoint (\#5) to viewpoint \#6 is shown in \cref{fig:generalize}. 
	
	Hence, the recorded kaleidoscopic image set yields the data samples as 262 single images, 2358 multi-view images, and 3144 disparity maps (\cref{Tab:dataset}). The disparity data for the corner images (\#1, \#3, \#7, \#9) include the displacements in two directions, \ie in $x$- and $y$-axis, with respect to the center viewpoint.
	\input{tabdataset}
	\subsubsection{Semi-automatic labeling}\label{sec:labeling}
	Our setup takes advantage of the multi-view data recording to capture 9x more data per image shot. This allows us to increase the available images with no extra cost or effort. In our PCB analysis framework, we define a total of 11 classes that need to be annotated on 2358 images. Obviously, with an increasing number of images and classes, and considering the difficulty of recognizing PCB components, annotation becomes increasingly complex and time-consuming. Thus, we develop a semi-automatic labeling technique for object detection to produce high-quality labeled data and efficiently save time and resources, as follows:
	
	\noindent- \textbf{Active labeling for single-view (center-view) images:} 1) A subset of center-view images (40\%) are manually annotated from scratch. 2) An object detector model is trained on these images. 3) The trained network is used for predicting the initial labels for the remaining center-view images. 4) The initial labels on center-view images are checked and modified.
	\input{figgeneralize}
	
	\noindent- \textbf{Depth-based label generalization for multi-view images:} With the help of disparity maps of the center viewpoint to the other viewpoints, each annotated bounding box in the center-view is translated in the proper direction along $x$- and/or $y$-axis. The value of the translation for each box is derived from the center of the bounding box in the disparity map. Therefore, in each kaleidoscopic image, the labels of the center-view are generalized to the other eight viewpoints using the corresponding disparity map. This means that for each image, we obtain the annotations of eight other images in a fully automatic manner. Namely, we define $B_{I_v}$ as the set of bounding boxes in viewpoint image $I_v$, with $v\in\{1,2,...,9\}$ as the viewpoint index:
	\begin{equation}
		\label{eg:1}
		B_{I_v}=\{b_1^{I_v},b_2^{I_v},...,b_n^{I_v}\}.
	\end{equation}
	Here, $n$ is the number of annotations (bounding boxes). The $i$-th bounding box in image $I_v$ is formulated as $b_i^{I_v}=(c_i^{I_v},x_i^{I_v},y_i^{I_v},w_i^{I_v},h_i^{I_v})$, where $(x_i^{I_v},y_i^{I_v})$ and $(w_i^{I_v},h_i^{I_v})$ indicate the center coordinates and the width/height of the bounding box, and $c_i^{I_v}$ denotes the ground-truth class category. We define a \emph{Forward Warping} function from center-view to viewpoint $v$, \ie $I_5 \rightarrow I_v$, in order to compute the \emph{disparity-displaced} bounding boxes:
	\begin{equation}
		\textrm{FW}_{B_{I_5}\rightarrow B_{I_v}}=\{b_1^{I_5 \rightarrow I_v},b_2^{I_5 \rightarrow I_v},...,b_n^{I_5 \rightarrow I_v}\}, \label{eq:FW}					
	\end{equation}
	\begin{equation}
		b_i^{I_5 \rightarrow I_v}=(c_i^{I_5},x_i^{I_5}-m, y_i^{I_5}-n,w_i^{I_5},h_i^{I_5}),
	\end{equation}
	\begin{equation}
		\begin{aligned}
			m = max(0,d_{{I_5}\rightarrow {I_v}}^X(x_i^{I_5},y_i^{I_5})), \\~n=max(0,d_{{I_5}\rightarrow {I_v}}^Y(x_i^{I_5},y_i^{I_5})).		
		\end{aligned}		
	\end{equation}
	Here, $m$ and $n$ indicate the amount of displacement in $x$ and $y$ directions, computed from the corresponding disparity map from center-view image to viewpoint $v$, \ie $d_{{I_5}\rightarrow {I_v}}^X$ and $d_{{I_5}\rightarrow {I_v}}^Y$. This process is summarized in \cref{fig:generalize}. It is important to note that the disparity maps, which link image contents from different perspectives, could make this approach feasible. Finally, the generated labels are checked to see if any components near the image borders are missed or need modification in case they are invisible in the center image. In this framework, by manually labeling only 105 images, we get the labels of 2253 other images automatically.
	\input{figComponents}
	\input{figcomponentstat}
	
	After analyzing the experimented boards, 11 classes are defined for their components: \emph{C} (Capacitor), \emph{D} (Diode), \emph{IC} (Integrated circuit), \emph{L} (Inductor), \emph{R} (Resistor), \emph{XTAL} (Crystals oscillator), \emph{LED} (Light-emitting diode), \emph{Q} (Transistor), \emph{AL\_C} (Aluminum capacitor), \emph{RY} (Relay), and \emph{FI} (Filter). An example of each component is shown in  \cref{fig:Components_}. Some classes, such as transistors, diodes, and inductors, could vary in appearance, shape, and text. Since aluminum capacitors are completely different from ceramic capacitors, an explicit class is assigned for them to enhance the network's learning capabilities. 
	\input{figalcr}
	
	\Cref{fig:component_stat} summarizes the statistics of various components after the labeling process in both single-view and multi-view images. \emph{C} and \emph{R} classes present more instances; others, with lower numbers, can be almost 9x greater in multi-view images. In  \cref{fig:al_c_r.png}, a sample component (aluminum capacitor) from different perspectives in multi-view images is illustrated.
	
	\subsection{Multi-view training}\label{sec:Multi-viewTraining}
	In order to identify components on PCBs, we use the object detection task, which involves 2D localization (defined by bounding boxes) and classification. To this end, we utilize the state-of-the-art model, YOLOv5 \cite{glenn_jocher_2020_4154370}, which has compound-scaled variants by introducing different numbers of layers and filters into its baseline architecture. We exploit the nano version since with only 1.9 million parameters, it suits a fast-operational industrial system. Also, two image dimensions, $640$ and $1024$ are considered for analysis. As part of YOLOv5, functions such as random HSV color space, left-right flipping, translation, and scaling are applied for online unique mosaic augmentation. When it comes to image augmentation, smaller models, \eg YOLOv5n, yield better results with less augmentation because overfitting can be avoided. However, with our hardware-assisted MV dataset, we can increase the number of samples safely with new real-world images.
	\subsection{Multi-view inference} \label{sec:mvi}
	This part introduces a novel approach for multi-view object detection, beyond training with multi-view data. Specifically, this technique applies in the \emph{inference} phase and requires \emph{multi-view test} data with a linkage between viewpoints spaces. The disparity map allows us to establish this connection. 
	
	To this end, we apply the trained model to each viewpoint image independently (center-view and other eight viewpoints), obtaining a total of nine object detection outputs. Probability scores and classification correctness can differ depending on the viewpoint. We leverage the ensemble of information from each viewpoint to reach a consensus from multi-view data, as follows: 
	\begin{itemize}
		\item \emph{Label warping}: When the model is applied to nine multi-view images, it generates an ensemble of boxes that need to be warped and aligned into a single view. To achieve this, we warp all bounding boxes to the center-view images. The disparity maps allow us to align the bounding boxes with the center-view. This is the reverse of the process explained for semi-automatic labeling in \cref{sec:labeling}. Instead of generalizing labels from the center-view to other viewpoints using disparity maps, we warp the predicted labels from other viewpoints to the center-view. Thus, we can define a \emph{Backward Warping} function similar to \eqref{eq:FW} that applies to each of the eight viewpoints surrounding the center-view.
		\item \emph{Bounding boxes fusion}: Following the warping, each object is accompanied by several predictions, which should be merged for the final prediction. A method such as Non-Maximum Suppression (NMS) \cite{neubeck2006efficient} can accomplish this by keeping only the bounding box with the highest Jaccard similarity score. Due to the strict suppression of other boxes by NMS, it is not suitable for our use-case, as we intend to systematically combine information from different predictions. A more advanced version of NMS is Non-Maximum Weighted (NMW) \cite{ning2017inception}, which is based on a weighted averaging of boxes to predict the average box, rather than the highest. In \cite{solovyev2021weighted}, authors proposed a similar strategy to fuse the ensemble boxes called Weighted Boxes Fusion (WBF) and proved that WBF outperforms NMW by using the confidence scores of the predicated bounding boxes to construct average boxes and comparing against the average box that is updated at each step of the comparison. 
		\item \emph{Intermixing the center-view coordinates}: Since the center-view coordinates are well-encompassing for components near the image border, we intermix them with the results from bounding boxes fusion. Namely, the class category and confidence score come from bounding boxes fusion, and the spatial coordinates from the center-viewpoint boxes. This way, we define the intermixing function as follows:
		\begin{equation}
			\textrm{IM}(b_i^{bbf}, b_i^{I_5}) = (c_i^{bbf},x_i^{I_5},y_i^{I_5},w_i^{I_5},h_i^{I_5},s_i^{bbf}),
		\end{equation}
		where $b_i^{bbf}$, $b_i^{I_5}$ represent the $i$-th bounding box from bounding boxes fusion and center-view image, respectively.
	\end{itemize}
	\section{Experimental results and discussion}
	\input{figsvvsmv}
	For training and validation, we conduct a $k$-fold cross-validation manner to robustly validate the performance measures. More precisely, we performed a 10-fold random split on the center-view (single-view) images, which we call the single-view (SV) dataset. After splitting the samples, the remaining viewpoints sets are added to the corresponding set (either the training set or the test set) to build the multi-view dataset (MV). As a result, we prevent viewpoint-related bias by assuring that all the viewpoints of a kaleidoscopic shot belong to either the training or the validation set.
	
	Evaluation metrics include precision, recall, mAP@0.5, and mAP@0.5:0.95 averaged over 10-fold cross-validation. mAP \cite{cocoapi} is most effective metric for evaluating object detectors.
	
	\noindent- \textbf{Multi-view training:} \Cref{tab:mv67} compares single-view (center-view) training with multi-view training (9x more training data) on two image sizes to demonstrate the performance of multi-view training. Single-view (SV) training is analogous to regular imaging without multi-view technology which we consider as the baseline model. By training the model with 9x more data, it safely achieves higher values in all metrics. This can be attributed to the enriched information of objects and to the background information, which is highly beneficial for object detection. Namely, the increased number of instances for each component class in the MV dataset is crucial in training a detection model, as it provides more varied and diverse examples of the components, enabling the model to learn more discriminative features for accurate detection. MV model also provides more robust results when images are downscaled. That is, the drop in accuracy of MV due to image size reduction is less than the case in SV model, which is beneficial for faster predictions or in limited-resources edge devices.
	
	\Cref{fig:sv_vs_mv} shows a sample result of SV and MV models and the activation maps of the predictions. In order to gain a better understanding of the model's feature representation, we used EigenCAM \cite{muhammad2020eigen} for displaying activation maps. Activation maps are a visualization technique used in deep learning to identify the most discriminative regions of an image that contribute to a network's behavior and its decision-making process. Based on the regions highlighted in the activation maps generated from the SV and MV models, it appears that MV model exhibits more activation than the SV model. This suggests that the model with higher activation may be more sensitive to certain image features, which leads to obtaining higher accuracy. 
	
	\noindent- \textbf{Multi-view inference:} For each center-view image, the other viewpoints are known for applying the MVI method. More specifically, in MVI, we assume multi-view test data is available; a model is trained on multi-view data; the model is tested on all nine images for a test sample; and finally, the results are fused. In \cref{tab:mv67}, we observe that multi-view inference achieve superior performance over SV and MV. In particular, using the advanced WBF, we can gain further $3.14$ accuracy in terms of mAP@0.5:0.95 compared to the vanilla MV model for image size 416$\times$640. 
	
	\Cref{tab:mv67component} presents the average results for each component individually in terms of mAP@0.5:0.95. The table validates the increase in performance by MV and MVI in all the component types. Notably, the improvement is especially significant for components that are difficult to detect. For example for XTAL and LED components in 672$\times$1024 image size, mAP@0.5:0.95 increases from 27.82\% to 86.01\% and from 59.08\% to 90.50\%, respectively.
	
	\input{tabmv67}
	\input{tabmv67component}
	\section{Conclusion}
	We presented a multi-view framework for analyzing PCB components. Our method works end-to-end, rather than extracting component regions first and then classifying them. The performance improves when multi-view images are used instead of single-view images. Furthermore, ensemble results of multi-view data are aggregated in the inference phase. We also developed a semi-automated labeling process to save the effort of manual labeling and ensure consistency of annotations for the object detection task.
	
	\section*{Acknowledgment}
	The authors would like to thank POLARIXPARTNER GmbH, \url{https://www.polarixpartner.com}, for providing the PCBs and their component data.
		
	{\small
		\bibliographystyle{ieee_fullname}
		\bibliography{egbib}
	}
	
\end{document}

%% file: figBlockgiagram.tex
\begin{figure*}[htbp]
	\centerline{\includegraphics[width=0.85\linewidth]{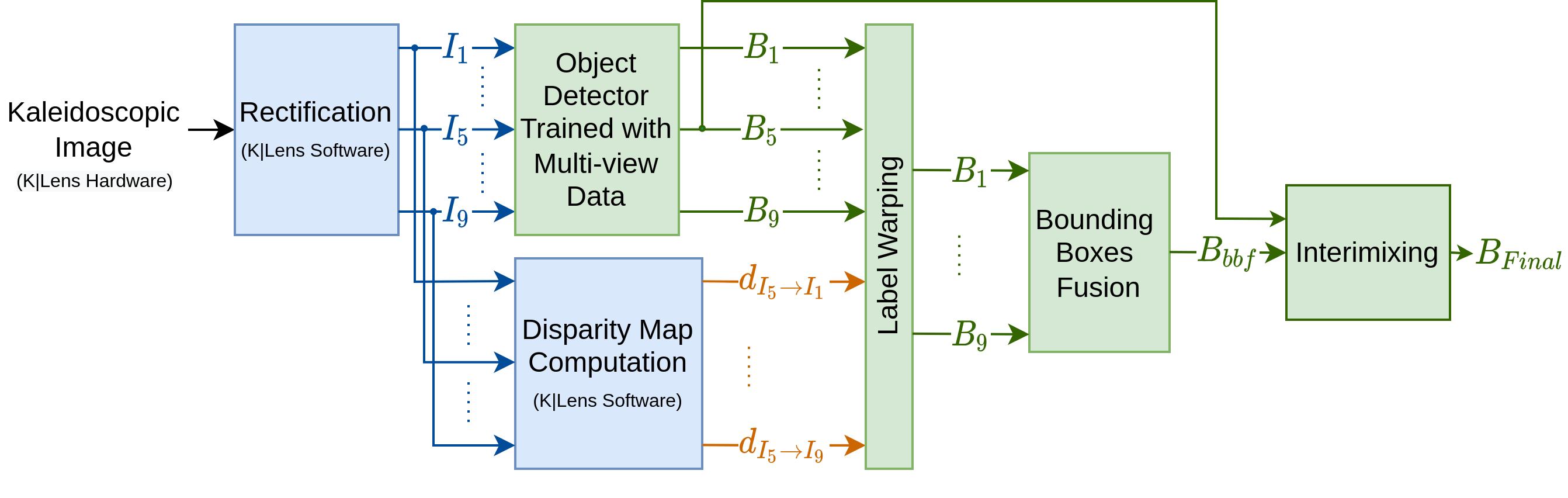}}
	\caption{Block diagram of the proposed component localization and identification using multi-view data. $I_i$ denotes the input images from different viewpoints and $I_5$ indicates the center viewpoint image. $B_i$ is the set of predicted results (bounding boxes) from the $i$-th image. An object detector is applied on each viewpoint independently following a warping function on the results using disparity maps. After aligning the results from the center viewpoint, the ensemble of predictions are aggregated together.}
	\label{fig:Blockgiagram}
\end{figure*}

%% file: fighardwaresetup.tex
\begin{figure*}[htbp]
	\centering
	\hspace{-2cm}
	\begin{subfigure}{0.33\linewidth}
		\centering
		\includegraphics[height=0.7\linewidth]{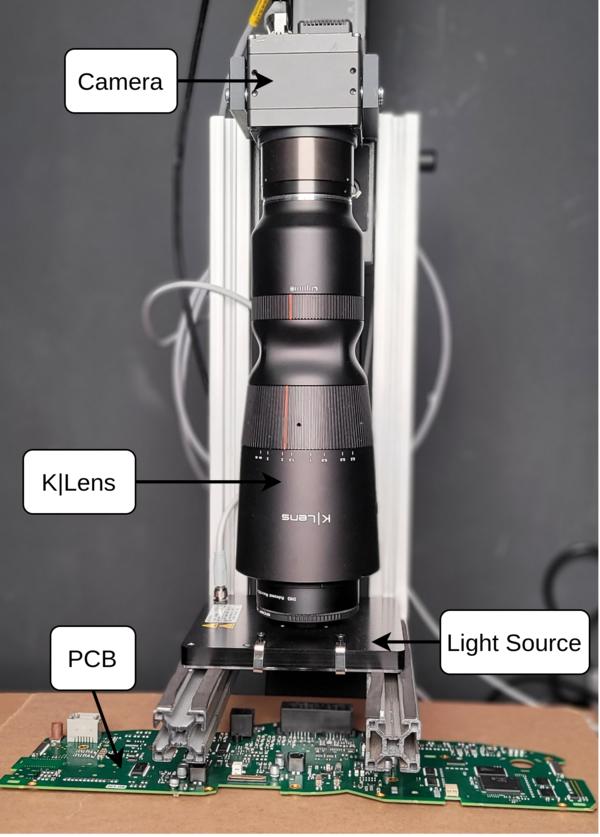}
		\caption{}
		\label{fig:a}
	\end{subfigure}
	\hspace{-1cm}
	\begin{subfigure}{0.33\linewidth}
		\centering
		\includegraphics[height=0.7\linewidth]{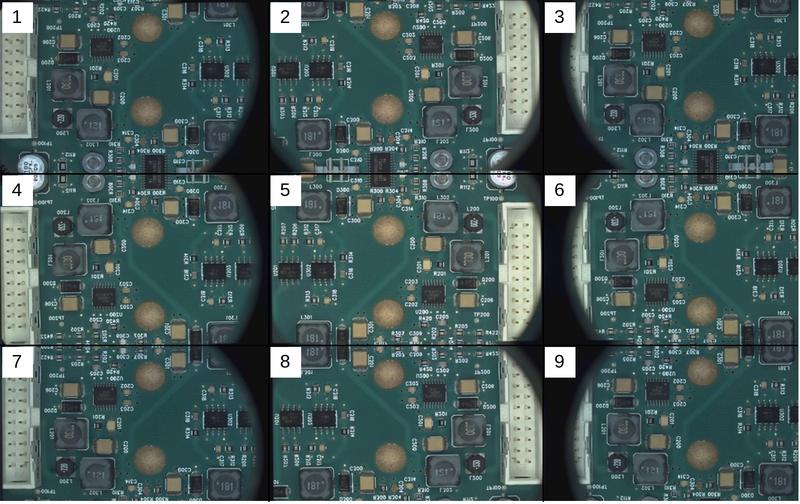}
		\caption{}
		\label{fig:b}
	\end{subfigure}
	\hspace{0.9cm}
	\begin{subfigure}{0.33\linewidth}
		\centering
		\includegraphics[height=0.7\linewidth]{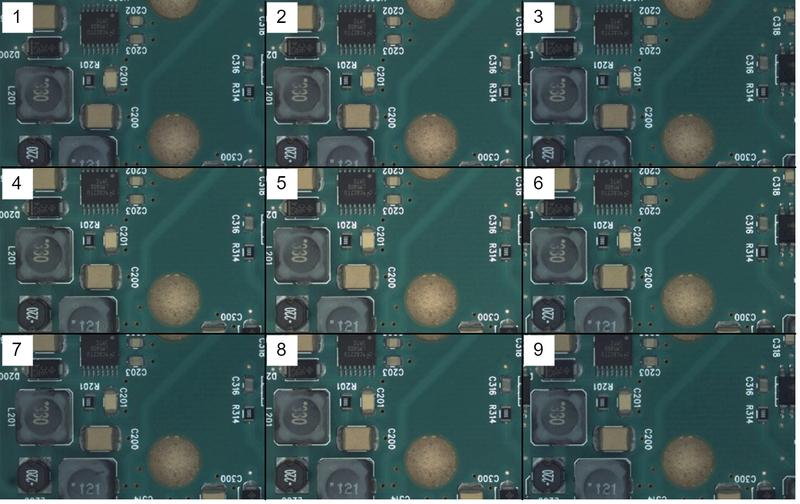}
		\caption{}
		\label{fig:c}
	\end{subfigure}
	\caption{(a) Our hardware setup for PCB analysis; (b) Sample kaleidoscopic image ($3\times3$ grid multi-view images); (c) Processed multi-view kaleidoscopic image after mirroring and rectifying the individual images and cutting out the distorted regions. Note that the individual images are piled up for the sake of comparison to the original kaleidoscopic image.}
	\label{fig:hardware_setup}
\end{figure*}

%% file: tabdataset.tex
\begin{table}[htbp]
	\centering	
		\begin{tabular}{@{}ccc@{}}
			\toprule
			\textbf{Single-view}	& \textbf{Multi-view (9x)}	& \textbf{Disparity maps} \\
			\midrule
			262 & 2358	 & 3144	 \\
			\bottomrule
		\end{tabular}
	\caption{K\textbar Lens PCB dataset collected from five PCBs using multi-view imaging.}
	\label{Tab:dataset}
\end{table}

%% file: figgeneralize.tex
\begin{figure}[htbp]
	\centerline{
		\includegraphics[width=1\linewidth]{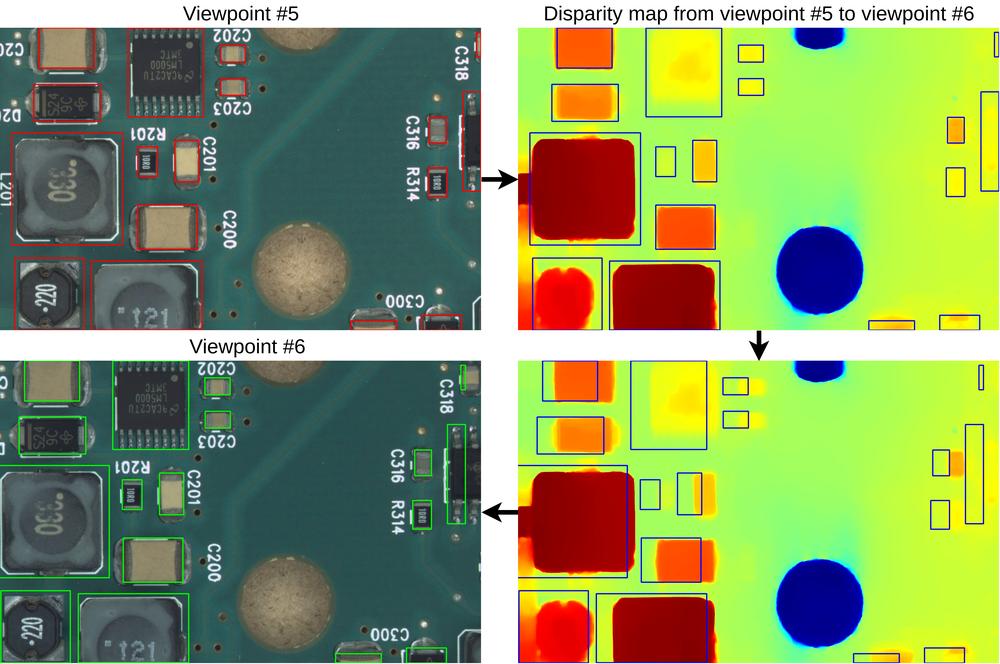}
	}
	\caption{Label Generalization: The disparity map from center viewpoint \#5 to viewpoint \#6 is generated by the system software. Each bounding box in the viewpoint \#5 is translated by the disparity map value in the center of the box. The translated labels align well with viewpoint \#6. }
	\label{fig:generalize}
\end{figure}

%% file: figComponents.tex
\begin{figure}
	\centering
	\includegraphics[width=1\linewidth]{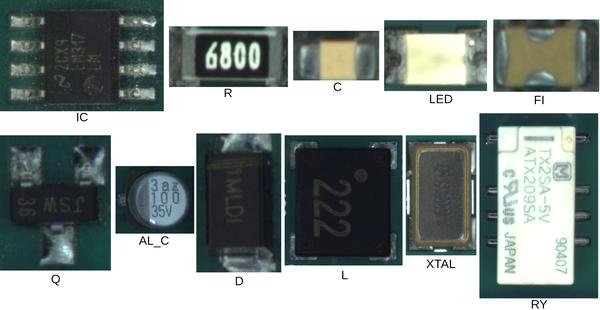}
	\caption{Samples of 11 component types for localization and classification. The following abbreviations are used for these classes: C (Capacitor), D (Diode), IC (Integrated circuit), L (Inductor), R (Resistor), XTAL (Crystals oscillator), LED (Light-emitting diode), Q (Transistor), AL\_C (Aluminum capacitor), RY (Relay), and FI (Filter). }
	\label{fig:Components_}
\end{figure}

%% file: figcomponentstat.tex
\begin{figure}[htbp]
	\centerline{
		\includegraphics[width=0.7\linewidth]{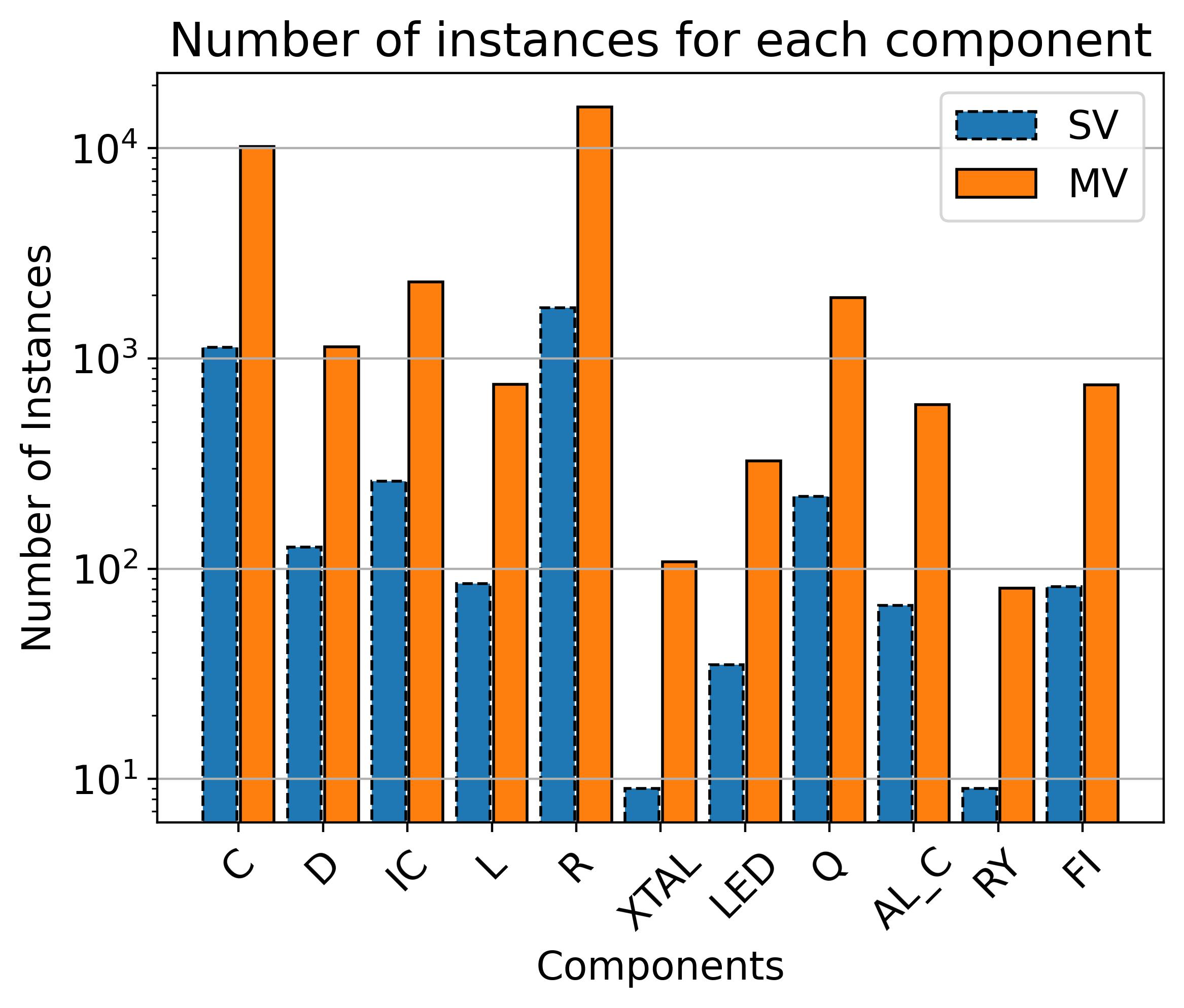}
	}
	\caption{Number of instances in defined classes for single-view and multi-view images. A logarithmic scale is used in the $y$-axis.}
	\label{fig:component_stat}
\end{figure}

%% file: figalcr.tex
\begin{figure}[htbp]
	\centering
	\includegraphics[width=0.7\linewidth]{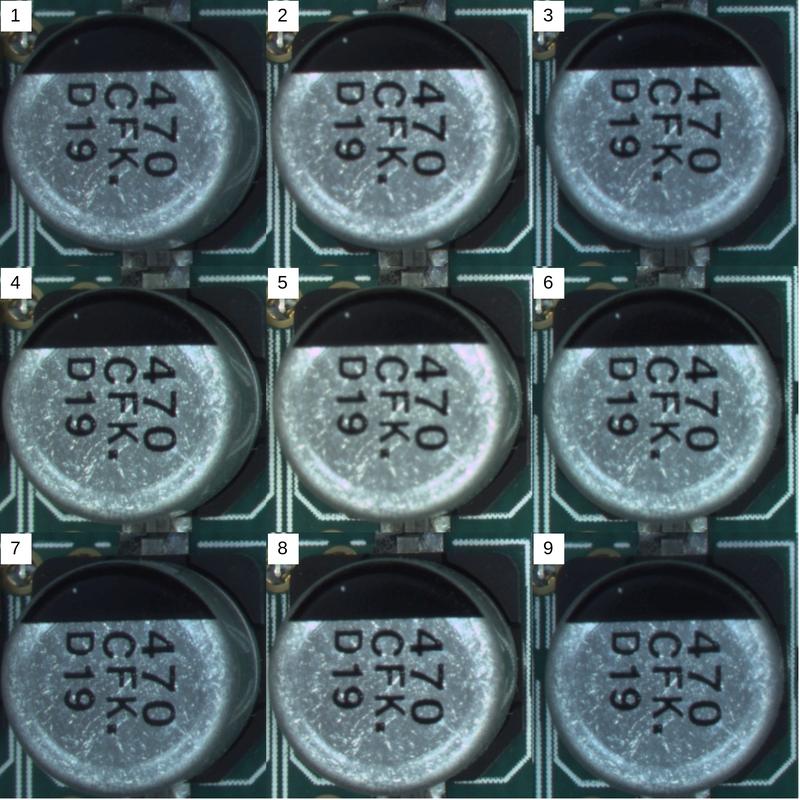}
	\caption{An annotated aluminum capacitor from different perspectives in multi-view data. Bounding boxes of the eight surrounding viewpoints are obtained via the proposed depth-based label generalization process.}
	\label{fig:al_c_r.png}
\end{figure} 

%% file: figsvvsmv.tex
\begin{figure*}[htbp]
	\centering
	SV\hspace{9cm} MV \\
	\includegraphics[width=0.495\linewidth]{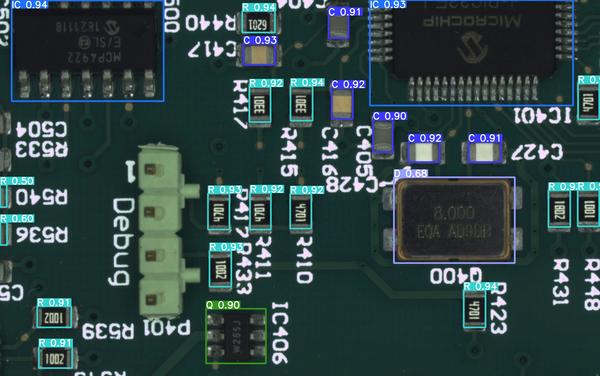}
	\includegraphics[width=0.495\linewidth]{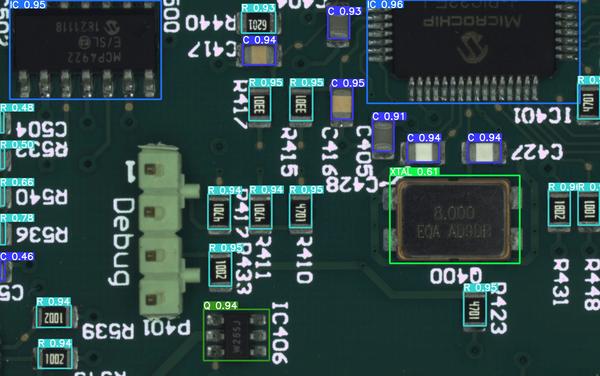} \\
	\includegraphics[width=0.495\linewidth]{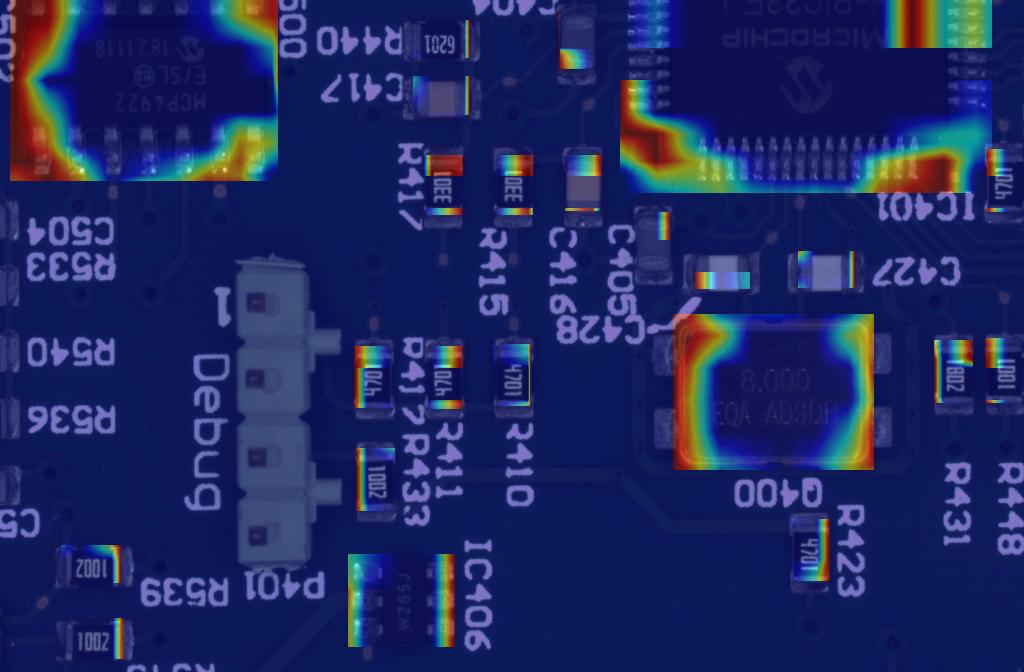}
	\includegraphics[width=0.495\linewidth]{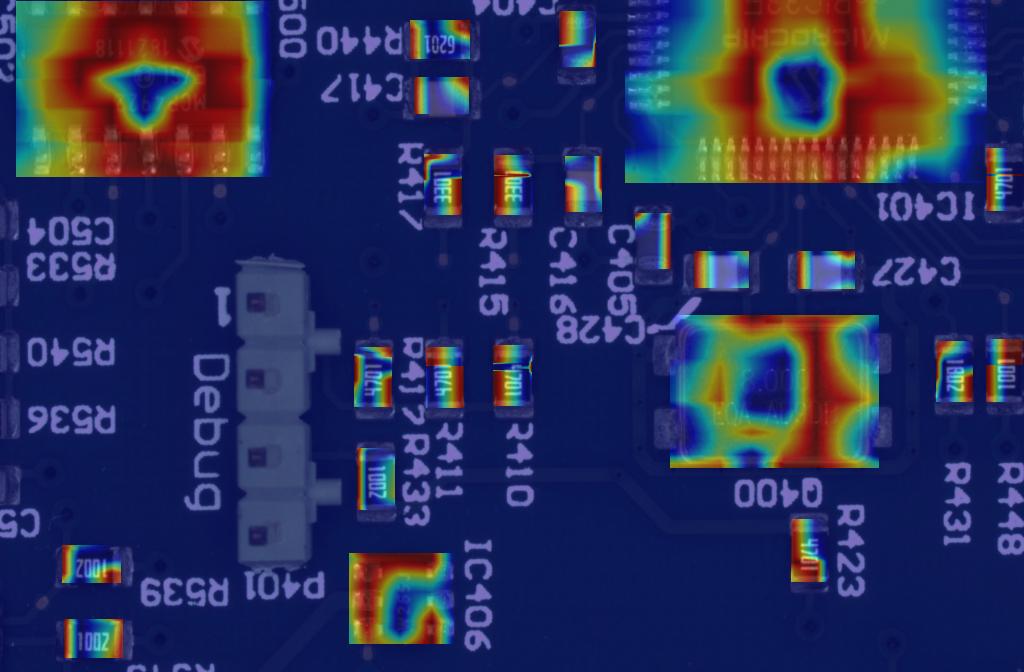}
	\caption{Qualitative performance using SV (\emph{Left}) and MV (\emph{Right}) models. The top row and bottom row respectively display the detection results and the activation maps. MV model exhibits more activation than the SV model, which leads to obtaining more accurate predictions.}
	\label{fig:sv_vs_mv}
\end{figure*}

%% file: tabmv67.tex
\begin{table}
	\centering	
	\resizebox{\linewidth}{!}{
		\setlength\tabcolsep{1pt}
		\begin{tabular}{@{}clcccc@{}}
			\hline
			\textbf{Image Size}   & \textbf{Model} & \textbf{Precision}$\uparrow$	& \textbf{Recall}$\uparrow$	& \textbf{mAP@0.5}$\uparrow$ & \textbf{mAP@0.5:0.95}$\uparrow$\\
			\hline
			\multirow{2}{*}{416$\times$640}  &SV (\textit{Baseline})			&  89.80			&	78.70				&	85.19				&	73.29	 			\\
			&MV			&  94.29&	91.46				&	94.86				&	85.87	 			\\
			&MVI		& \textbf{94.44}	&	\textbf{93.35}	&	\textbf{96.59}		&	\textbf{89.01}	 	 \\
			\hline
			\multirow{2}{*}{672$\times$1024}  	&SV	(\textit{Baseline})		& 94.58				&	79.12				&	87.46				&	77.41 				\\
			&MV			& 94.79&	92.23				&	96.02				&	88.81 				\\
			&MVI	& \textbf{95.55}	&	\textbf{93.64}	&	\textbf{97.17}		&	\textbf{91.04} 		\\
			\hline
		\end{tabular}
	}
	\caption{Comparison of PCB component detection performance using single-view (SV), multi-view (MV) and multi-view inference (MVI) approaches, based on the average metrics using 10-fold cross-validation.}
	\label{tab:mv67}
\end{table}

%% file: tabmv67component.tex
\begin{table}
	\centering
		\begin{tabular}{@{}lcccccc@{}}
			\toprule
			\multirow{2}{*}{\textbf{Class}} & \multicolumn{3}{c}{\textbf{416$\times$640}} & \multicolumn{3}{c}{\textbf{672$\times$1024}} \\
			\cline{2-7}
			&  SV	& MV	& MVI	&  SV	& MV	& MVI \\
			\midrule
			C		&86.32	&91.28	&\textbf{92.05}	&91.57	&\textbf{95.73}	&95.52		\\
			D 		&72.28	&80.87	&\textbf{83.71}	&81.87	&85.46	&\textbf{88.98}		\\
			IC		&91.53	&96.22	&\textbf{97.41}	&94.03	&96.54	&\textbf{97.27}		\\
			L		&81.87	&84.24	&\textbf{87.54}	&82.00	&86.25	&\textbf{89.48}		\\
			R 		&87.47	&92.72	&\textbf{93.43}	&92.00	&\textbf{95.59}	&95.36		\\
			XTAL	&31.79	&81.53	&\textbf{89.61}	&27.82	&74.77	&\textbf{86.01}		\\
			LED		&33.23	&81.82	&\textbf{85.44}	&59.08	&89.43	&\textbf{90.50}		\\
			Q 		&81.59	&87.54	&\textbf{91.15}	&85.92	&90.25	&\textbf{92.86}		\\
			AL\_C	&87.51	&90.88	&\textbf{92.35}	&89.07	&92.35	&\textbf{93.70}		\\
			RY		&66.48	&68.99	&\textbf{75.65}	&60.67	&74.49	&7\textbf{6.10}		\\
			FI 		&86.22	&88.37	&\textbf{90.80}	&87.59	&\textbf{96.15}	&95.68		\\
			\bottomrule
		\end{tabular}
	\caption{Performance evaluation of PCB component detection on each component individually in terms of mAP@0.5:0.95 using 10-fold cross-validation on center-view test data.}
	\label{tab:mv67component}
\end{table}